\pdfoutput=1
\RequirePackage{fix-cm}
\documentclass[smallextended]{svjour3}       
\smartqed  
\usepackage{graphicx}
\usepackage{amsmath}
\usepackage{marvosym}
\usepackage{array}
\newcolumntype{C}[1]{>{\centering\arraybackslash }b{#1}}

\usepackage{color}
\newcommand{\modified}[1]{\textcolor{black}{#1}}

\begin{document}

\title{Real-time Analysis of Cataract Surgery Videos using Statistical Models
}


\author{Katia~Charri\`ere,
        Gw\'enol\'e~Quellec,
        Mathieu~Lamard,
        David~Martiano,
        Guy~Cazuguel,
        Gouenou~Coatrieux
        and~B\'eatrice~Cochener 
}

\authorrunning{Katia Charri\`ere et al.} 

\institute{K. Charri\`ere (\Letter), G. Coatrieux and G. Cazuguel \at
INSTITUT MINES-TELECOM; TELECOM Bretagne; UEB; Dpt ITI, Brest, F-29200 France \\
\email{ katia.charriere@telecom-bretagne.eu.}\and
All authors \at
LaTIM - INSERM UMR 1101, Brest, F-29200 France\and
M. Lamard and B. Cochener \at
Univ Bretagne Occidentale, Brest, F-29200 France \and
B. Cochener and D. Martiano \at
CHRU Brest, Service d'Ophtalmologie, Brest, F-29200 France
}

\date{Received: date / Accepted: date}

\maketitle

\begin{abstract}
The automatic analysis of the surgical process, from videos recorded during surgeries, could be very useful to surgeons, both for training and for acquiring new techniques. The training process could be optimized by automatically providing some targeted recommendations or warnings, similar to the expert surgeon's guidance. In this paper, we propose to reuse videos recorded and stored during cataract surgeries to perform the analysis. 
The proposed system allows to automatically recognize, in real time, what the surgeon is doing: what surgical phase or, more precisely, what surgical step he or she is performing. This recognition relies on the inference of a multilevel statistical model which uses 1) the conditional relations between levels of description (steps and phases) and 2) the temporal relations among steps and among phases. The model accepts two types of inputs: 1) the presence of surgical tools, manually provided by the surgeons, or 2) motion in videos, automatically analyzed through the Content Based Video retrieval (CBVR) paradigm. Different data-driven statistical models are evaluated in this paper. 
For this project, a dataset of 30 cataract surgery videos was collected at Brest University hospital. The system was evaluated in terms of area under the ROC curve. Promising results were obtained using either the presence of surgical tools ($ A_z = 0.983 $) or motion analysis ($ A_z = 0.759 $).
The generality of the method allows to adapt it to any kinds of surgeries. The proposed solution could be used in a computer assisted surgery tool to support surgeons during the surgery.

\end{abstract}

\section{Introduction}
%
%
%
%

Training acquisition of new techniques is a very important part of a surgeon's career and it requires a major investment from expert surgeons in terms of supervision. The work presented in this paper aims to support the training process by automatically analysing the surgical process in surgery monitoring videos. This analysis could be used in the future for providing targeted recommendations or warnings, similar to the guidance of expert surgeons.
 
Our first goal is to support surgeons during their first surgeries and reduce the time of expert supervision. Assisting the surgeon during the surgery implies to be able to analyze the surgical process in real time. This could be done through the automated analysis of the recorded video. Also, information recorded during previous surgeries could be reused to recognize similar situations during the analysis of a new surgery video. 
The methods we developed could also take their place with the emergence of surgery simulators \cite{singh2014high} to support the initial training of surgeons, even before their first (supervised) surgery. In this scenario, rather than the video stream, we could use the presence (or trajectories) of surgical tools to perform the analysis.

Although this problem is common to any surgery, we focus in this paper on eye surgeries and the cataract surgery in particular. Cataract surgery aims to replace the eye's natural lens by a synthetic lens when its transparency has been lost. This is one of the most practiced surgery. In this surgery, the surgeon watches the patient's eye through a binocular microscope, the output of which can be video recorded.
An accurate analysis of the surgery is necessary to be able to provide relevant information. Ideally, we would like to know, at each instant of the surgery, which surgical gesture or step is being performed by the surgeon. But an accurate analysis of the surgical process, with the real-time constraint, is a challenging task: the algorithm has to be fast and we have to be able to recognize a quite large number of surgical steps.

\begin{figure*}[!t]   
   \centering
   \includegraphics[scale=0.49]{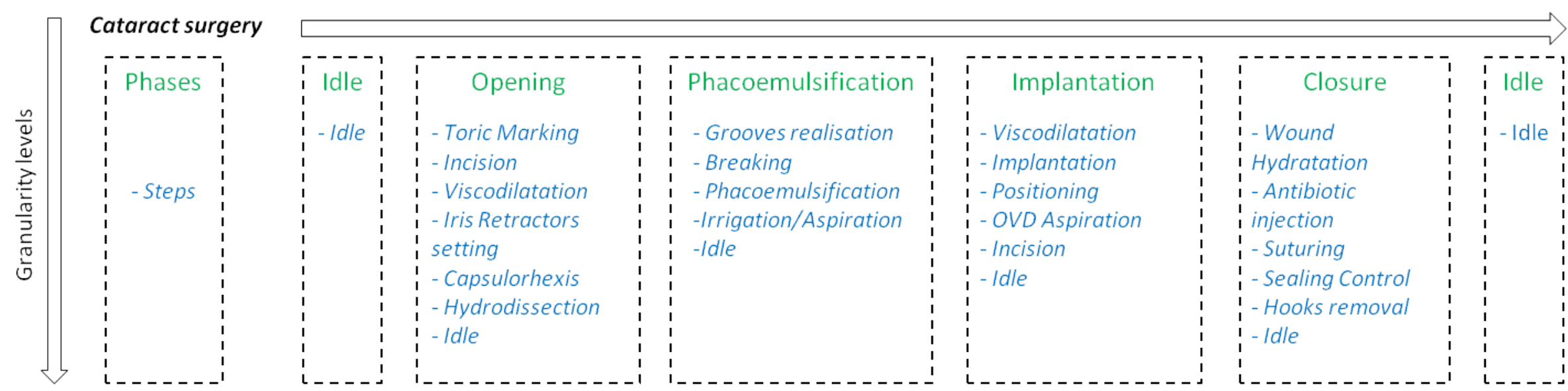} 
   \caption{Description of the cataract surgery into surgical phases and surgical steps}
   \label{fig:fig1}  
\end{figure*}

In this paper, we propose to work with two levels of description, in order to perform a robust and precise analysis of the surgical process. As presented in Fig.~\ref{fig:fig1}, at the coarsest level, the surgery is divided into "phases" and at the finest level, it is divided into "steps". By learning a multilevel statistical model of the surgical process, the proposed methods use both the knowledge of the temporal process of the surgery and the knowledge of the relationships between the surgical steps and phases. The system provides the most probable sequences of surgical steps and phases, given present and past information only, by assigning a label for steps and phases to each image of the video. In the field of medical data analysis, an ongoing problem is the small number of available data. Building an annotated medical dataset is a challenging task, first because of the sensitivity of the data. Also, the annotation of data by experts is time consuming and surgeons' time is precious. Our system needs to deal with this limitation of data, especially to train a model of the surgical process. Our system also needs to deal with the specifity of the medical data recorded from video monitored surgery. In the case of the cataract surgery, the system has to deal with the motion of the eye during the surgery, and some zoom or level variations.

 The remainder of this paper is organized as follows. Related work is presented in \ref{sec:RelWork}. The proposed methods are described in \ref{sec:Methodology}. In \ref{sec:Exp}, we discuss our experimental setup and results. Finally, \ref{sec:Concl} presents some concluding remarks.

%
\section{Related work} 
\label{sec:RelWork}

To analyze the surgical process, a surgery can be conveniently regarded as a succession of surgical gestures (finest representation), activities, steps, tasks or phases (coarsest representation). Every surgical process analysis method describes the surgery at a given level of abstraction. A high level of description, into surgical phases \cite{lalys2012framework} or tasks \cite{Quellec2014b}, provides a global description of the surgery with a simple sequencing. Indeed, surgeries generally have the same phase sequencing \cite{padoy2012statistical,lalys2013automatic}.
Automatic recognition at a low level of description, on the other hand, is a challenging task because of the large number of possible temporal sequences, sometimes marginally represented in the dataset. So, a low level of description, into surgical gestures \cite{zappella2013surgical}, activities \cite{lalys2013automatic} or steps, allows a more precise analysis of the surgery, but implies a more complex surgical process.

Some authors propose to work with several levels of description. For instance, Padoy et al. \cite{padoy2012statistical} use the presence of the surgical tools in the field of view of the camera to detect actions. This action detection supports the recognition of (high-level) phases. Forestier et al. \cite{forestier2015automatic} use low-level recordings of the activities that are performed by a surgeon to automatically predict the current phase of the surgery. Those two methods allow an on-line recognition of (high-level) phases, but low level information (presence of surgical tools or activities) is manually provided by surgeons. Lalys et al. propose a method based on an automated extraction of the visual content of the video and use the sequential nature of the surgical phases as a temporal constraint for (low-level) activity detection \cite{lalys2013automatic}.

In terms of methodology, several methods reuse data recorded during a video-monitored surgery for the automated analysis of surgical processes \cite{lalys2013automatic,zappella2013surgical}. In particular, some of these methods rely on Content-based video retrieval (CBVR), whose goal is to find similar videos or sub-videos inside a dataset \cite{chattopadhyay2008application,Quellec2014c,quellec2015real,loukas2016shot}. But those methods do not model the temporal sequencing of the surgical process. Different kinds of models were used to model this process, like Dynamic Time Warping (DTW) averaging \cite{padoy2012statistical,lalys2013automatic}, which builds an average surgery. But this method does not allow on-line computations because it requires the entire video to be known (past, present, but also future information). On the other hand, Hidden Markov Models (HMMs) \cite{padoy2012statistical} or their derivatives, like Conditional Random Fields (CRFs), do allow on-line computations. CRFs seems to provide better results than the HMMs in the context of automatic surgical video analysis \cite{Quellec2014b,tao2013surgical}. A Hierarchical Hidden Markov Model (HHMM), a hierarchical generalization of HMMs, was also used by \cite{twinanda2016endonet} to perform a phase recognition taking into account inter-phase and intra-phase dependencies.

A majority of methods for the analysis of video data  are developed without the real-time constraint: they are applied to automatic documentation and report generation \cite{stanek2012automatic,lalys2013automatic}, fast search of similar cases in a database \cite{andre2012learning} or educative video construction \cite{cao2008medical}.
A few of them allow on-line analysis \cite{padoy2012statistical,forestier2015automatic}, but at a high level of description (into surgical phases) and do not allow accurate analysis of the surgery. Lalys et al. \cite{lalys2013automatic} propose a finer analysis of the surgery, but this method is not able to perform on-line analysis of the surgical process.

The method presented in this paper extends a previous solution from our group, presented at a conference \cite{charriere2016real}. 
That system performs an on-line analysis of a cataract surgery video at two different levels of description. It uses high-level phase recognition to help low-level step recognition, but it also uses information from step recognition to refine the recognition of phases.

\section{Methodology} 
\label{sec:Methodology}

In this paper, we present a comprehensive study comparing several surgical process models, working at multiple description levels, and using various kinds of observations from the video stream. Those models aim to model the temporal process at each level of description as well as the relationships between steps and phases. 

First a Hierarchical Hidden Markov Model (HHMM) is evaluated. The advantage of HHMM is that is can jointly model the temporal process at multiple description levels, using a simple relationship model between levels. We then evaluate two novel models which handle separately the relation relationships between steps and phases and the temporal process.
A first model is composed of a Bayesian network (BN) and Hidden Markov Models (HMMs). The BN is used to model the conditional relationships between steps and phases while the temporal relationships at each level of description are modeled by HMMs.  We also evaluate a variation on this model where HMMs are replaced by Conditional Random Fields (CRFs). 
These models can take as inputs different kinds of observations, computed throughout the surgery. These observations are presented in the following section. Next, the models are presented. 

\subsection{Observations}
\begin{figure}[!h]   
   \centering
   \includegraphics[scale=0.7]{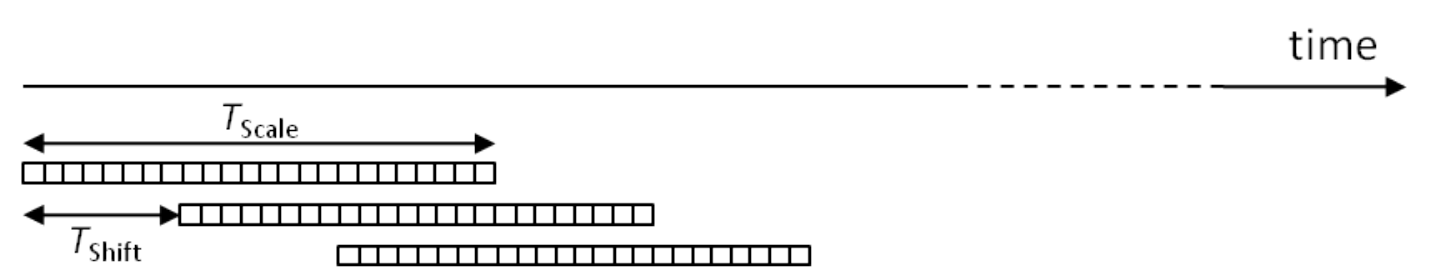} 
   \caption{The video divided into overlapping fixed-size sub-sequences}
   \label{fig:fig2}  
\end{figure}

To generate observations, the video is divided into overlapping fixed-sized sub-sequences. Each sub-sequence contains the same number of frames, as presented in Fig.~\ref{fig:fig2}. The best length of sub-sequences ($T_{\text{scale}}$) and the best temporal shift ($T_{\text{shift}}$) between them are chosen after a learning step (\S~\ref{Training_eval_proc}). For each sub-sequence, observations are computed. In this paper, observations derive either from the presence of surgical tools or from the analysis of motion in videos through the content-based video retrieval (CBVR) paradigm. In the case of the motion analysis, we evaluate the model using Motion Histograms (MH) \cite{Quellec2014c} as feature vectors, but also using Bags of Visual Words (BoVW) \cite{wang:inria-00439769,zappella2013surgical}. As motion extraction can be disturbed by scale variation or motion of the eye, we also evaluate the system after a spatial normalization of images from the video stream.

\subsubsection{Presence of surgical tools}
We can assume than the use of a specific tool is closely related to the surgical steps and phases being performed. In a scenario where the system is used jointly with a surgery simulator, we can easily assume that this information will be provided by the simulator. In a scenario where the system is used during a real surgery, this information may also be obtained using barcodes or RFID chips \cite{roberts2006radio,yao2010rfid}. Another solution consists in automatically tracking surgical tools in the surgical scene by computer vision methods. In this paper, this information is manually provided by surgeons, which allows us to validate the models.

\subsubsection{Video motion comparison}
\label{sec:videoMotionComparison}

Because the presence of tools cannot always be obtained easily, we also try to evaluate the model using general motion features extracted from the video content. 
In this paper, we compare Motion Histograms (MH), based on the extraction of the optical flow, with the widely used Bag-of-Visual-Words (BoVW) model. Disruptive motion could appear, induced by camera or eye motion during cataract surgeries for exemple. The influence of a spatial normalization of images that composed the video on the performance of the system is evaluated. This normalization step aims to refine feature extraction based on pupil center and scale tracking \cite{charriere2014automated}.  

\begin{itemize}
\item Normalization:
Pupil center and scale tracking are obtained without explicitly segmenting these landmarks, as presented on \cite{quellec2014normalizing}. First, a robust solution to track the pupil center is used: it uses the fact that the pupil boundaries, the limbs and the sclera / lid interface are concentric. Then, the zoom level is estimated from the illumination pattern reflected on the cornea. Knowing this information for each frame of the video, we can pre-processe them to balance eye motion, zoom or level variation before the feature extraction step. First, eye motions are balanced by registering all frames on the same iris center. 
A simple coordinate system change is applied, which places the iris center at the image center. This should allow eliminate motion induced by eye or camera motion and make tool motion more relevant. Then, all frames are scaled on a same scale level to balance zoom or level variations. 
After this last preprocessing step, all irises should have the same radius. Finally, a circular mask centered on the iris center is applied to select a region of interest, because all relevant actions should appear in a region closed to the iris location. 
 
\item Motion Histograms (MH): 
To compute the optical flow, strong corners are first detected and selected. The OpenCV 2 library\footnote{http://opencv.org/} is used to select strong corners and the optical flow between each pair of consecutive frames is computed at each strong corner by the Lucas-Kanade iterative method \cite{lucas1981iterative}. Finally, the motion contained in the sub-sequence as a whole is characterized by one 8-bin amplitude histogram, two 8-bin amplitude-weighted spatial histograms (one for the x-coordinates and one for the y-coordinates) and one 8-bin amplitude-weighted directional histogram.
\item Bags of Visual Words (BoVW): 
The BoVW features are based on Space-Time Interest Points (STIP), which were proposed by Laptev et al. \cite{laptev2005space}. STIP points are first detected locally within each sub-sequence. Histograms of oriented gradient (HOG) and histograms of optical flows (HOF) are then extracted inside a cube centered around each STIP point and concatenated. During a learning step, those local feature vectors are used to build a dictionary of visual words. Once the dictionary is learnt, a histogram of visual words is extracted from each video sub-sequence and used as a feature vector for the sub-sequence as a whole.

\end{itemize}

Finally, given a feature vector (MH or BoVW), the $K$ nearest neighbors of each sub-sequence are found in the training set by comparing sub-sequences with a Euclidean distance. The best number of nearest neighbors ($K$) is also chosen after a learning step. For each sub-sequence, the labels of the nearest sub-sequences at the finest granularity level (steps), provide the probability of belonging to each step. 

This concludes our presentation of observations. The statistical models are presented next.

\subsection{Statistical Models of the surgical process}

Let $ \mathcal{O} = \{o_{t}\}_{t=1}^{T}$ be a sequence of observations, where $o_{t}$ represent the observation generated at time $ t $. Let $ \mathcal{S}=\{ (s_{i})_{0<i\leq N_{\text{s}}} \} $ be the $ N_{\text{s}} $ labels for steps and $ \mathcal{P}=\{ (p_{j})_{0<j\leq N_{\text{p}}} \}$ be the $ N_{\text{p}} $ labels for phases. Our goal is, given a model, to find the sequences of labels for steps and phases, denote respectively by $ \mathcal{Y}_{steps} = \{y_{t}^{steps}\}_{t=1}^{T} $ and $\mathcal{Y}_{phases} = \{y_{t}^{phases}\}_{t=1}^{T} $, that are most likely to generate the observation sequence. 

The evaluated multilevel statistical models try to represent both the relationships between steps and phases and the temporal relationships at each level of description. For instance, if an "Incision" step is being performed, the probability of an "Opening" phase is high. Conversely, if an "Opening" phase is being performed, the probability of a "Stitching up" step is very low. Also, if an "Incision" step is being performed by the surgeons, we can refine the probabilities by knowing that a "Stitching up" step has a really low probability of occurrence in the next sub-sequence. The different surgical steps and phases identified by the surgeons of the Brest hospital are presented on Fig.~\ref{fig:fig1}.

\subsubsection{HHMM}
\begin{figure}[!h]   .
   \centering
   \includegraphics[scale=0.4]{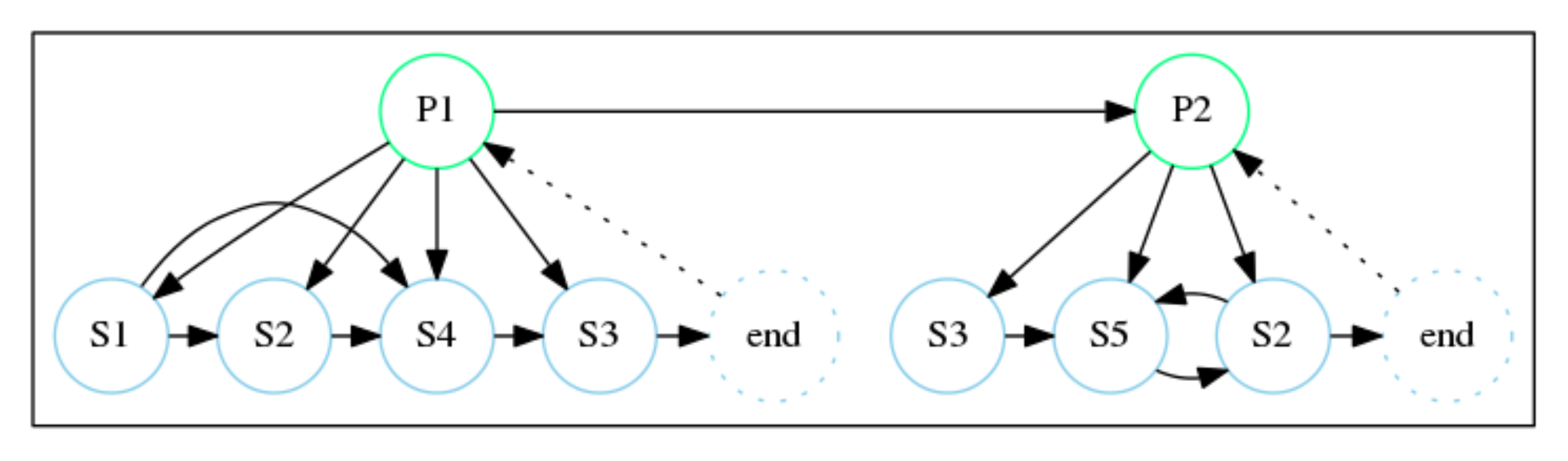} 
   \caption{Example of HHMM; green nodes represent labels for phases and blue nodes represent labels for steps}
   \label{fig:fig3}  
\end{figure}

First, a Hierarchical Hidden Markov Model (HHMM) is used to model the relationships between steps and phases and the temporal relationships. The HHMM derives from the HMM and each state of the HHMM is an HHMM as well. In our HMMM, each state denoted by $q_ {i} ^ {d} $ represent a step or phase label. Each state produces a sequence of symbols (instead of a unique symbol), through a process of recursive activations which ends when a production state is reached. The production states (as opposed to the internal states) are the only states which emit observable symbols.  
In our case, labels for phases are represented by some internal state, and each state for phase is itself a HHMM composed by some production states which represent the labels for steps (Fig.~\ref{fig:fig3})

Following the notations introduced by \cite{fine1998hierarchical}, a state of a HHMM is denoted by $q_ {i} ^ {d} ~ (d \in \{1,...,D\})$ where $q_ {i} ^ {d} \in \{\mathcal{S},\mathcal{P}\}$ and $i$ represent the state index and $d$ the hierarchical index. In our case, $D = 2$. A HHMM is defined by the following set of parameters: \modified{$\{\{A^{q^{d}}\}_{d\in\{1,...,D-1\}},$ $\{\Pi^{q^{d}}\}_{d\in\{1,...,D-1\}},\{B^{q^{D}}\}\}$}. For each internal state a transition probability matrix is associated, denoted by $A^ {q^ {d}} = a_ {ij} ^ {q^ {d}} $. The transition probabilities \modified{$a_{ij}^{q^{d}} = P(q_{j}^{d+1}|$ $q_{i}^{d+1})$} represent the probability of making a horizontal transition between the $i$th to the $j$th substate of $q^{d}$.
The matrix $\Pi^{q^{d}}=\{\pi^{q^{d}}(q_{i}^{d+1})\}=\{P(q_{i}^{d+1}|q^{d})\}$ represent the initial probability of the substates of $q^{d}$. This can also be interpreted as the probability of making a vertical transition, which is the probability of entering substate $q_{i}^{d+1}$ from $q^{d}$. Then, an output probability vector $B^{q^{D}}$ is also associated with each production state. 

To link the presence of the surgical tool with the model, we compute the output probability vector $B^ {q^ {D}} $, the probability of being on a step $s_ {I} \in \mathcal{S}$ from a given pair of surgical tools. In the case of motion analysis as observation (\S \ref{sec:videoMotionComparison}), probabilities provided by the KNN search directly provide the conditional probability \modified{$ P( state | obs. )$} required for the inference. All probabilities are learned by frequency counting in the training set.

In HHMM, the relationships between steps and phases are modeled very simply: we only indicate which steps can happen in each phases. In particular, we do not indicate the probability of coocurrence between a step and a phase. In this paper, we propose to model these cooccurence relationships using a Bayesian network (BN). Once these relationships are modeled, temporal relationships can be analyzed independently for steps and for phases. With this relaxation, multiple temporal models can be used, such as hidden Markov models (HMMs) and conditional random fields (CRFs).

\subsubsection{Bayesian networks to model the cooccurrence of steps and phases}

Bayesian networks are convenient as they can model the influence of step
occurrence on phases but also the influence of the phase
occurrence on steps, Bayesian networks are suitable models \cite{pearl1998bayesian}. Each label for steps and phases is represented by a node in the network. Each conditional relationship between a step $ s_{i} \in \mathcal{S} $ and a phase $ p_{j} \in \mathcal{P} $ is represented by an edge $ (s_{i},p_{j}) $. As we need to link the observations with the model, some observation nodes are added. We denote by $ O=\{ (o_{k})_{0<k\leq N_{\text{o}}} \} $ the $ N_{\text{o}} $ observation nodes. 
Thus, we define the structure of the Bayesian network by the graph $ \mathcal{G} = (\mathcal{V},\mathcal{E}) $ where $ \mathcal{V}= \{ V_ {l} \}_{0<l\leq(N_{\text{s}} + N_{\text{p}} + N_{\text{o}})} $ represents the sets of nodes and $ \mathcal{E} $ represents the set of edges. As we need to link the observations, obtained from the visual content of the video, with the model, some observation nodes are added. If the presence of surgical tools in the field of view of the camera is available, each surgical tool is associated with an observation node: the observation is true if and only if the tool is present. If we consider motion analysis, probabilities defined in \ref{sec:videoMotionComparison} need to be converted to Boolean evidence (true or false), for compatibility with Bayesian networks: each observation node is associated with a range of probabilities (for instance: "the KNN search indicates a probability between 10\% and 20\% that "Incision" is being performed in the sub-sequence").
Thus, we define the structure of the Bayesian network by the graph $ \mathcal{G} = (\mathcal{V},\mathcal{E}) $ where $ \mathcal{V}= \{ V_ {l} \}_{0<l\leq(N_{\text{s}} + N_{\text{p}} + N_{\text{o}})} $ represent the sets of nodes and $ \mathcal{E} $ represent the set of edges. Conditional probabilities associated with each edge are learned by frequency counting in the training set.

\subsubsection{ BN + HMM}
\begin{figure}[!h]   
   \centering
   \includegraphics[scale=0.5]{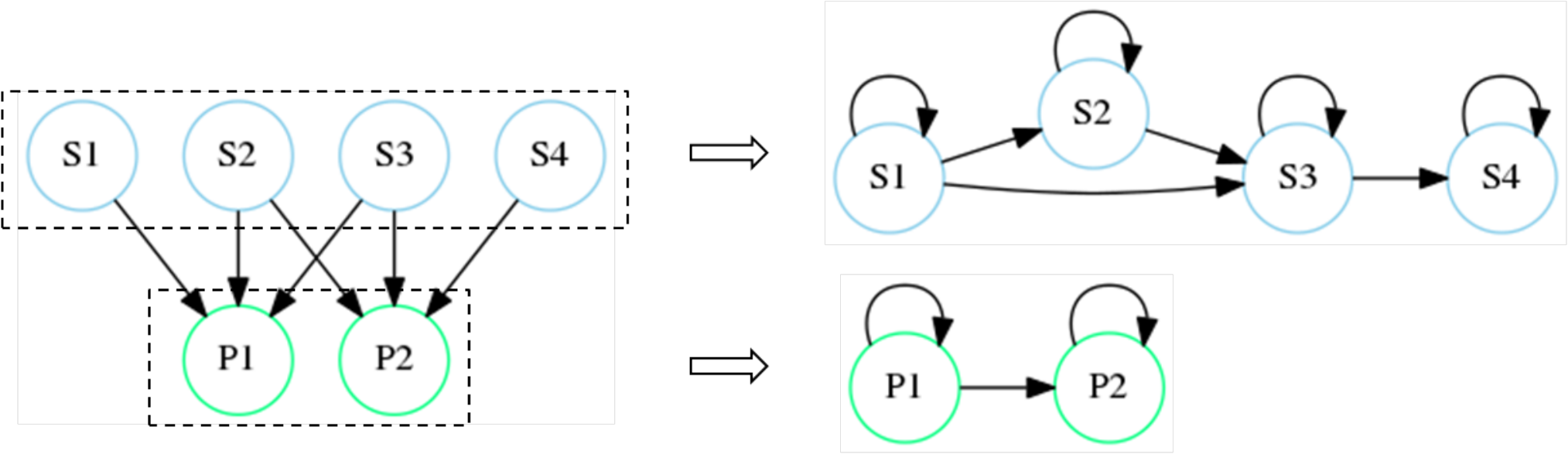} 
   \caption{Example of a BN (on the left) with two HMMs (on the right); green nodes represent labels for phases and blue nodes represent labels for steps}
   \label{fig:fig4}  
\end{figure}

One HMM is defined for each granularity level: one for the step level, called $ hmmS $, and one for the phase level, called $ hmmP $. A HMM is defined by a quadruplet $ \{ S, \Pi_{i}, A, B \} $. The sets of states $ S_{steps} = \mathcal{S} $ and $ S_{phases} = \mathcal{P} $ are defined respectively by the $ N_{\text{s}} $ labels for steps and the $ N_{\text{p}} $ labels for phases. The transition probabilities $ a_{ij} = P ( q_{t} = s_{j} | q_{t-1} = s_{i}) $ which compose the transition matrices are obtained by counting transitions in the training set: $ a_{ij} = \frac{N_{ij}}{\sum_{j}N_{ij}} $, where $ N_{ij} $ represent the number of cases in the training set where we observe a transition from state $ S_{i} $ to state $ S_{j} $. In our case, it is not necessary to define the observation probability matrices $ B_{steps} $ and $ B_{phases} $, because the conditional probability \modified{$ P( state | obs. )$} is given by the Bayesian network inference (Fig.~\ref{fig:fig4}).

\subsubsection{Extension}
We also evaluate an extension of this model, for which we reuse the output probabilities for phases as additional observation in the BN. This aims to add a major influence of phase recognition (which should be easier) on step recognition. 
The results of HMM for phases for the previous subsequence are added as complementary observation. In this case, some observation nodes are added to the models and linked with nodes represented phase labels. Each observation node is associated with a range of probabilities.

\subsubsection{BN + CRFs}
A limitation of HMMs is that, if the training set does not represent all possible transitions, some transitions will likely not be recognized during the labelling session. CRFs are trained without explicitly counting cooccurrence frequencies in a reference dataset. Also, they do not consider consecutive steps only. Therefore, they are less limited by dataset size \cite{lafferty2001conditional}.

Following the previous model, a Bayesian network is used to model the conditional relationships between the occurrence of steps and phases. But, the two HMMs are replaced by two CRF : one CRF for each granularity level. In a CRF model, the conditional probability of the label sequence $\mathcal{Y}$, given the observation sequence $\mathcal{O}$ could be modelled by the equation:
\begin{equation}
\label{eq:one}
\left\{
    \begin{aligned}
P(\mathcal{Y} | \mathcal{O})  & \propto \exp(\sum_{t=1}^{T}\mathbf{\lambda}_{t}\Psi_{t}^u(y_{t},o_{t}) \\
        		& + \sum_{t=1}^{T-1}\mathbf{\mu}_{t}\Psi_{t,t+1}^p(y_{t},y_{t+1},o_{t}))
    \end{aligned}
\right.
\end{equation}

where $\Psi_{t}^u$ and $\Psi_{t,t+1}^p$ are the CRF unary and pairwise potentials.  $ \mathbf{\lambda} $ represent a vector of weights over the unary potentials and $ \mathbf{\mu} $ a vector of weights over the pairwise potentials learnt from training data by maximum log-likelihood estimation. 

The unary potentials represent the score of assigning a label to an observation. In our case, the definition of potentials can only rely on the present and past information. Those potentials depend on observations and are obtained with the Bayesian network which computes, at each time $t$, the probabilities of occurrence for each step and phase $P(l_{i}|o_{t})$ where $ l_{k} \in \mathcal{S}$ or $\mathcal{P}$. The unary potentials are defined as follows:  
\begin{equation}
\label{eq:two}
\Psi_{t}^u(y_{t},o_{t}) = log(P(l_{i}|o_{t-\Delta_{t}})), 0\leq\Delta_{t}< t  
\end{equation}

The pairwise potential represents the probability of switching from label $ l_ {i} $ to $ l_ {j} $ when moving from an observation to another one. The relationship between two adjacent observations is given by the transition probability $P (l_ {j} |l_ {i})$ computed from the training dataset. The pairwise potentials are defined as follows:
\begin{equation}
\label{eq:three}
\Psi_{t,t+1}^p(y_{t},y_{t+1},o_{t}) = log(P(l_{j}|l_{i})) 
\end{equation}

The Wapiti library\footnote{https://wapiti.limsi.fr/} is used for training the CRF. The \modified{L-BFGS\footnote{\modified{Limited-memory Broyden-Fletcher-Goldfarb-Shanno}}} Quasi-Newton algorithm is used to learn the weight vectors $ \mathbf{\lambda} $ and $ \mathbf{\mu} $.

\begin{figure*}[!t]   
   \centering
   \includegraphics[scale=0.48]{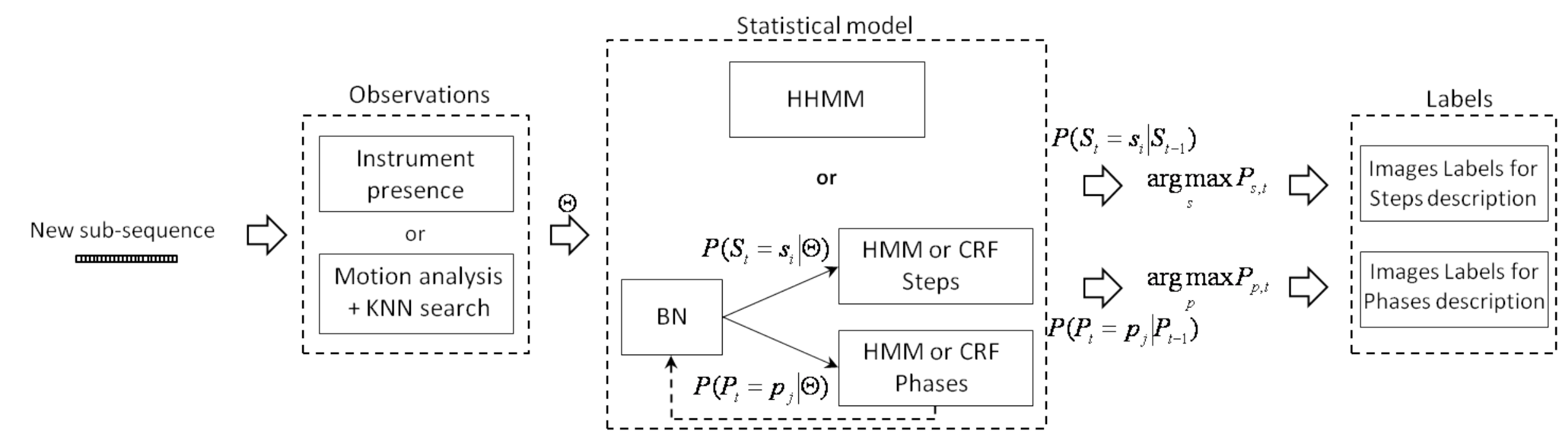} 
   \caption{Methodology of the system}
   \label{fig:fig5}  
\end{figure*}

\subsection{On-line recognition}

Each constructed model is used to determine which labels for steps and phases are associated with each subsequence of the video given the observation extracted from that subsequence. Inference algorithms are described below.

\subsubsection{HHMM inference}

Inference of the HHMM is based on the generalized Viterbi algorithm proposed by \cite{fine1998hierarchical}, adapted to an on-line constraint. For the presence of the surgical tools as evidence, the production states use this Boolean evidence. For the motion analysis the probabilities \modified{ $P( state | obs. )$} are directly provided by the KNN search to the production states. At the production state level, the inference is similar to the Viterbi algorithm \cite{forney1973viterbi}. When a final state at the step level (production states) is reached, this implies a transition a the phase level (internal states), as follows:
\begin{itemize}

\item Initialization: 

$ \delta(t,t,q_{i}^{d-1},q^{d-1}) =\pi^{q^{d-1}}\delta(t,t,q_{r}^{d+1},q^{d-1}).a_{r~end}^{q_{i}^{d}} $
\item At each time step: 

$ 
    \begin{aligned} 
\delta&(t,t+k,q_{i}^{d-1},q^{d-1}) 
 \\
        		& = \max_{1 \leq j \leq |q^{d-1}|} [\delta(t,t+k-1,q_{j}^{d-1},q^{d-1}).a_{ji}.(\delta(t,t,q_{r}^{d+1},q^{d-1}).a_{r~end}^{q_{i}^{d}})] 
    \end{aligned}
$

\end{itemize}
The transition at the phase level, activates the HHMM (at the step level) associated with the new phase. This inference finds the most likely multilevel state sequence.

\subsubsection{BN + HMMs inference}

The Bayesian network determines a probability of occurrence for each phase and step label, given the observations. Those probabilities are then used during the HMMs inferences to determine the most likely labels for steps and phases.

First, Bayesian network inference is performed thanks to the D-lib library\footnote{http://dlib.net/} using a method from the MCMC (Markov chain Monte Carlo) family of methods: the Gibbs sampler algorithm. This inference algorithm performs the computation of posterior probabilities, given the new observations, which are provided to the Bayesian network through observation nodes. 
The inference algorithm  computes the $P(l_{i} | o_{t}) $ probabilities at each temporal step $ T_{\text{hmmS}} $ and $ T_{\text{hmmP}} $, where $l_{i} \in \mathcal{S}$ or $\mathcal{P}$. Then, those probabilities are used by the two HMMs. The HMMs inference is performed with the Viterbi algorithm \cite{forney1973viterbi}. 
This inference finds the most likely sequence of hidden states at each granularity level. 

\subsubsection{BN + CRF inference}

Similar to the previous model, the BN provide probabilities of each labels for steps and phases to the CRF. CRF inference is usually performed with the forward-backward algorithm \cite{sha2003shallow}, but in our case we would like to perform an on-line analysis, that is during the execution of the surgery. So we can only use information from past and present. The forward algorithm determines the probability $P_{t}(l_{i})$ to obtain label $l_{i}$ for the sub-sequence recorded at time t, where $l_{i} \in \mathcal{S}$ or $\mathcal{P}$ \cite{Quellec2014b} as follows:

\begin{equation}
\label{eq:for}
\left\{
    \begin{aligned}
        \log(P_{0}(l_{i})) & = \sum_{u=1}^{U}\lambda_{u}\Psi_{0}^{u}(l_{i},o_{t}) \\              
        \log(P_{t}(l_{i})) & = \max_{l_{i}\in \mathcal{S} or \mathcal{P}}						[log(P_{t-1}(l_{j})) + \sum_{u=1}^{U}\lambda_{u}\Psi_{t}^{u}(l_{i},o_{t}) \\
        		& + \sum_{b=1}^{P}\mu_{b}\psi_{t-1,t}^p (l_{j},l_{i},o_{t}) ]
    \end{aligned}
\right.
\end{equation}
The Wapiti library was used for CRFs inference, with a modification to allow online inference.

\begin{table*}[!t]
\renewcommand{\arraystretch}{1.3}
\caption{Evaluation of the model consisting of a Bayesian network and two HMMs with the different sources of observations as input. $A_ {z} $ is the area under the ROC curve and the last row represents the number of frames we are able to process in a second.}
\label{table_one}
\centering
\begin{tabular}{|c||C{0.88cm}|C{0.88cm}|C{0.88cm}|C{1cm}|C{1cm}|C{0.88cm}|C{1cm}|}
\hline
    & \multicolumn{5}{c|}{Motion analysis}  & \multicolumn{2}{c|}{Tools}  \\
\cline{2-8}
    & MH & BoVW  & MH + Norm. & MH + feedback HMM~P & MH + Norm. + feedback HMM~P & tools  & tools + feedback HMM~P \\
\hline
\hline    
$A_{z}$ Steps & 0.674 & 0.686 & 0.721 & 0.676 & \textbf{0.733} & 0.903 & \textbf{0.946} \\
$A_{z}$ Phases & 0.812 & 0.828 & \textbf{0.832} & 0.811 & 0.819 & \textbf{0.922}  & 0.910 \\
\hline
$A_{z}$ Means & 0.743 & 0.757 & 0.777 & 0.743 & \textbf{0.779} & 0.913 & \textbf{0.928}  \\
\hline
Nr Frames / s & 13.2 & 0.92 & 11.22 & \textbf{14.2} & 9.88 &\textbf{21.5}  & 16.9 \\
\hline
\end{tabular}
\end{table*}

The statistical models provide the probability occurrence of each surgical step and phases. The label with the maximum occurrence probability (for steps and phases) are associated with the images of the $T_{\text{shift}} $ interval.

\section{Experiments}
\label{sec:Exp}

After a presentation of the dataset collected for this study, the influence of the different sources of observations as input of the models and the influence of the statistical models are evaluated. 

\subsection{Cataract Surgery Database}

A database of 30 cataract surgeries was used in this study. Surgeries were performed by different surgeons. Videos were recorded in DV format. This database was manually annotated by surgeons using a description in phases and steps. Surgeons also manually indicated timing for the presence of surgical tools in the camera's field of view. Five phases and 20 steps were identified by surgeons, as presented in Fig.~\ref{fig:fig1}. Surgeries in the database were not always performed in the same way. Surgeries are surgeon- and patient-dependent. The incision step, for instance, is strongly dependent of surgeons. The "Phacoemulsification" step has a variable execution time depending on the stage of development of cataract.  Finally, the "Closure" phase could be realized by several "Wound Hydration" steps or by a "Point Suturing" step.

\subsection{Training and evaluation procedure}
\label{Training_eval_proc}
As we need a sufficient number of examples to build the statistical model, and as our database is quite small, the evaluation of the system was performed through a 6-fold cross validation. The database was partitioned into six subsets of five videos. Each subset was used as test set, while the other subsets were used as training set. The training set was used to learn the model (structures and probabilities), and to optimize the parameters through a grid search. The parameters $T_{\text{shift}}$, $T_{\text{scale}}$, the HMMs' time steps $T_{\text{hmmS}}$, $T_{\text{hmmP}}$ or the number of potentials for the CRFs are only evaluated for the model using the presence of surgical tools as observation, which is the most reliable: the same parameters were used in the model using motion analysis. 

We evaluated the performance of our system by measuring the area under the Receiver Operating Characteristic (ROC) curve for each step and phase. We built one ROC curve for each step and phase and for each fold. We computed the mean area $A_{z} $ for steps, phases and a mean of the two levels. We also evaluated the number of frames that the system is able to process in a second using one core of a « quad core » Intel(R) Core(TM) i7-3770 (3.40GHz) processor.

\begin{table*}[!t]
\renewcommand{\arraystretch}{1.3}
\caption{Evaluation of the three models with Motion Histograms (MH) and the presence of the surgical tools in the field of view of the camera (tools) as observations }
\label{table_two}
\centering
\begin{tabular}{|c||C{1.15cm}|C{1.15cm}|C{1.15cm}|C{1.15cm}|C{1.15cm}|C{1.15cm}|}
\hline
    & \multicolumn{3}{c|}{MH}  & \multicolumn{3}{c|}{Tools}  \\
\cline{2-7}
    & BN + HMMs & BN + CRFs  & HHMM & BN + HMMs & BN + CRFs  & HHMM \\
\hline
\hline    
$A_{z}$ Steps & 0.674 & 0.691 & 0.521 & 0.903 & \textbf{0.980} & 0.908 \\
$A_{z}$ Phases & 0.812 & 0.828 & 0.517 & 0.922  & \textbf{0.986} & 0.844 \\
\hline
$A_{z}$ Means & 0.743 & 0.759 & 0.520 & 0.913 & \textbf{0.983} & 0.863 \\
\hline
Nr Frames / s & 13.2 & 13.2 & 3507 & 21.5  & 21.7 & \textbf{4791} \\
\hline
\end{tabular}
\end{table*}

\subsection{First experiment: influence of observations}

First, the influence of the different observations on the performance of the system was evaluated on the model, composed by a BN and two HMMs. For the model using motion analysis, the number $K$ of nearest neighbors and the number of classes to represent output KNN probabilities by observation nodes $N_{\text{obs}}$ was also learned. After the learning step, ($T_{\text{scale}}$) and ($T_{\text{shift}}$) were set, respectively, to two seconds and one second for each validation set. 
The three kinds of input were evaluated and compared. For motion analysis, the two features (MH and BoVW) were first compared, then the influence of the spatial normalisation of images was evaluated with the MH as features. Then, the presence of the surgical tools as observations was evaluated as input of the model. We also evaluated a combination of the presence of the tools and the results of the HMM output for the previous subsequence as observations. The results are presented in Table \ref{table_one}.

The good results obtained using the presence of tools as observations confirm that the use of surgical tools is strongly correlated with step occurrence. And because the system was able to process about 21 frames per second using one core, it shows that the system is compatible with the real time constraint. Recognition performance was quite inferior for surgical steps, because at this level the surgical process is more complex, with a larger number of possible transitions. 
With the motion analysis as observations, the results were also inferior. That was expected because, in this case, the observations are automatically extracted from the visual content. BoVW features provided better performance in terms of $A_{z}$ than MH. But BoVW are not compatible with the real time constraint. Results with MH were satisfactory anyway, especially given the number of frames we were able to process in one second for an entirely automated system. The spatial normalization of images improves performance in terms of ROC curve with a mean area of $0.777$ instead of $0.743$, but the number of processed frames per second is slightly reduced. However, the system is always compatible with the real-time constraint.
When the presence of tools is used as observations, feeding back the results of the HMM for phases as complementary observation improves performance, especially for steps recognition. It shows that the recognition of the surgical phases also has an influence on the recognition of the surgical steps.

\subsection{Second experiment: model comparison}

We then compare the different statistical models with two kinds of observations: the motion analysis with Motion Histograms (MH) as features and the presence of the surgical tools in the field of view of the camera (tools) as observations. The results are presented in Table \ref{table_two}. 

The model composed by a BN and two CRFs achieved the best performances with a mean area under the ROC of up to 0.98 achieved for the two levels of description, with the presence of the surgical tools in the field of view of the camera. The improvement of the results obtained by replacing HMMs with CRFs can be explained by the fact that our training set does not represent all the possible transitions and CRFs better handle the small amount of examples. The HHMM is very fast, with more than 4500 images processed per second. But the results are inferior than results obtains with BN+CRFs or BN+HMMs and are very low with the motion histograms as features for the motion analysis. The HHMM inference is not able to detect transitions well, especially with a noisy input.

\section{Discussion and Conclusion}
\label{sec:Concl}

In this paper, we have proposed several systems, based on statistical models, able to analyze a cataract surgery in real time (during the execution of the surgery). The recommended system consists of a Bayesian network and two CRFs because of its ability to model the relationships between steps and phases as well as temporal relationships, and that with a low number of examples in the training set. The system is easily adaptable to other video-monitored surgeries. 

The proposed system works at two levels of description, it allow an accurate and complete analysis of the surgery. This is adapted for the generation of specific warnings and recommendations in order to supplement supervision by expert surgeons. Given a surgery, the systems provide the most likely sequences of surgical steps and phases.  The system has been evaluated with two levels of description, but they could be adapted easily to any number of levels. 

Also, the system allows to easily evaluate any kind of observations (or combinations of observations). Good results were obtained using the presence of surgical tools in the field of view of the camera as observation, at the input of the statistical model, especially when the model consisted of a Bayesian network and two CRFs (one for each level of description). In this configuration, the labeling step was almost perfect and a mean area under the ROC curve of 0.982 was achieved. The presence of tools could be obtained easily in the case of a surgical simulator. In the case of a use during a real surgery, it would be very interesting to develop a solution to automatically recognize the surgical tools. For a use during a real surgery instead of a simulator, an automated system could be developed to recognize the surgical tools and enhance the CBVR tools.
Indeed, more contrasted results were obtained with the automated generation of observations from the visual content of the video through motion analysis. Some good results were obtained for the recognition of surgical phases with a mean area under the ROC Curve of 0.828. But the results were quite inferior for the recognition of surgical steps with a mean area of 0.691. The results were improved by a spatial normalization of images that composed the videos.

The low number of examples in our dataset impacts the performance of recognition with models based on HMMs (HHMM or BN+HMMs) because of independences found after the training step. If a transition is not represented in the training set, it would not be recognized during the analysis of a surgery. But this problem is well handled by replacing HMMs by CRFs. 

In conclusion, a general framework was proposed for the automatic sequencing of surgeries and encouraging results were obtained in a dataset of cataract surgery videos.

\begin{acknowledgements}
The authors would like to thank the Urban Community of Brest (Brest M\'etropole Oc\'eane) and the "Institut Mines-T\'elecom" for funding this project.
\end{acknowledgements}

\bibliographystyle{spmpsci}      
\bibliography{biblio}   

\end{document}